\documentclass[11pt,a4paper]{article}
\usepackage[margin=1in]{geometry}
\usepackage{amsmath,amssymb,amsfonts,amsthm}
\usepackage{graphicx}
\usepackage{booktabs}
\usepackage{hyperref}
\usepackage{cleveref}
\usepackage{subcaption}
\usepackage{xcolor}
\usepackage{algorithm}
\usepackage{algpseudocode}
\usepackage{multirow}
\usepackage{float}

\title{Hallucination as Trajectory Commitment:\\Causal Evidence for Asymmetric Attractor Dynamics\\in Transformer Generation}

\author{
Gokturk Aytug Akarlar\thanks{Corresponding author. \texttt{akarlaraytu@gmail.com}. Chimera Research Initiative is an independent research effort exploring causal and neuro-symbolic approaches to AI systems.} \\
Chimera Research Initiative \\
Istanbul, T\"{u}rkiye
}

\date{April 2026}

\newtheorem{definition}{Definition}

\begin{document}
\maketitle

\vspace{-0.5em}
\begin{center}
\small Code and data: \url{https://github.com/akarlaraytu/trajectory-commitment}
\end{center}
\vspace{0.5em}

\begin{abstract}
We present causal evidence that hallucination in autoregressive language models is an \emph{early trajectory commitment} governed by asymmetric attractor dynamics. Using same-prompt bifurcation, in which we repeatedly sample identical inputs to observe spontaneous divergence, we isolate trajectory dynamics from prompt-level confounds. On Qwen2.5-1.5B across 61 prompts spanning six categories, 27 prompts (44.3\%) bifurcate with factual and hallucinated trajectories diverging at the first generated token ($\text{KL} = 0$ at step 0, $\text{KL} > 1.0$ at step 1). Activation patching across 28 layers reveals a pronounced causal asymmetry: injecting a hallucinated activation into a correct trajectory corrupts output in 87.5\% of trials (layer 20), while the reverse recovers only 33.3\% (layer 24); both exceed the 10.4\% baseline ($p = 0.025$) and 12.5\% random-patch control. Window patching shows correction requires sustained multi-step intervention, whereas corruption needs only a single perturbation. Probing the prompt encoding itself, step-0 residual states predict per-prompt hallucination rate at Pearson $r = 0.776$ at layer 15 ($p < 0.001$ against a 1000-permutation null); unsupervised clustering identifies five regime-like groups ($\eta^2 = 0.55$) whose saddle-adjacent cluster concentrates 12 of the 13 bifurcating false-premise prompts, indicating that the basin structure is organized around regime commitments fixed at prompt encoding. These findings characterize hallucination as a locally stable attractor basin: entry is probabilistic and rapid, exit demands coordinated intervention across layers and steps, and the relevant basins are selected by clusterable regimes already discernible at step~0.
\end{abstract}

\section{Introduction}

Large language models hallucinate; they generate plausible but factually incorrect text with high confidence \cite{ji2023survey,huang2023survey}. Despite extensive empirical study, the internal mechanisms governing hallucination remain poorly characterized. Existing work has established that hallucination correlates with identifiable features in model internals: probes over hidden states can detect hallucination above chance \cite{azaria2023internal,li2023inference}, entropy-based signals precede hallucinated outputs \cite{kadavath2022language}, and representation engineering can partially steer models toward truthfulness \cite{li2023inference,zou2023representation}.

Yet correlation does not establish mechanism. A central question persists: \emph{when and where does a model commit to a hallucinated trajectory, and is this commitment causally reversible?}

We address this question through two methodological contributions:

\paragraph{Same-prompt bifurcation.} Rather than comparing different prompts that elicit correct versus hallucinated outputs, which conflates prompt-level semantics with trajectory-level dynamics, we sample the \emph{same prompt} repeatedly under non-zero temperature and identify prompts where the model produces \emph{both} factual and hallucinated completions. This isolates the trajectory-level divergence: identical initial states yield different outcomes solely through the stochastic sampling path.

\paragraph{Symmetric causal patching.} For each bifurcating prompt, we collect correct and hallucinated runs with full hidden-state caches. We then perform bidirectional activation patching: replacing a hallucinated run's activation with a correct run's activation (and vice versa) at each layer and generation step, with three control conditions (random-prompt patch, wrong-to-wrong patch, unpatched baseline).

\begin{figure}[t]
    \centering
    \includegraphics[width=0.75\textwidth]{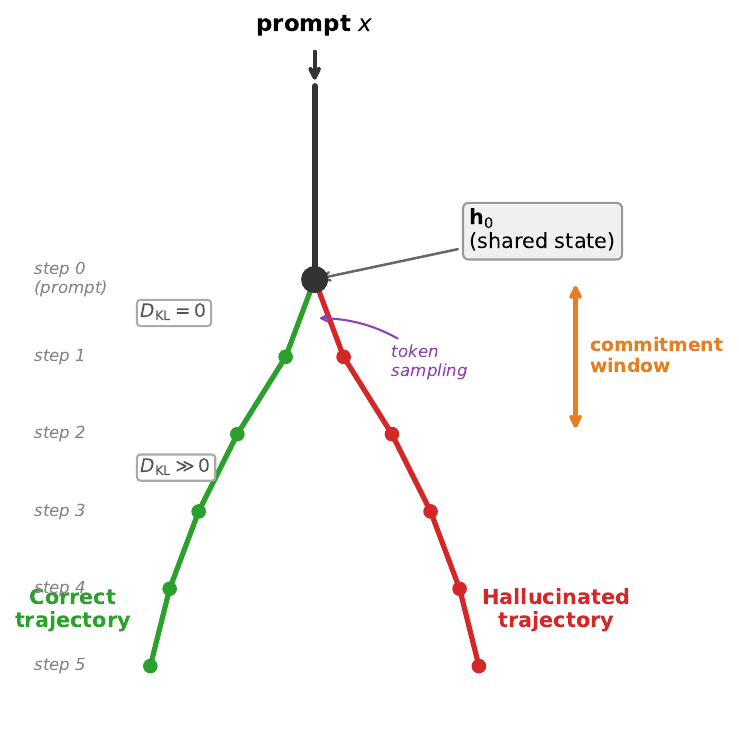}
    \caption{Conceptual overview. From a shared initial state $\mathbf{h}_0$, stochastic token sampling commits the trajectory to either a correct (green) or hallucinated (red) basin. Activation patching reveals that corruption (crossing into the hallucination basin) requires only a single-point perturbation, while correction (escaping the hallucination basin) requires sustained multi-step intervention, the hallmark of an asymmetric attractor landscape.}
    \label{fig:conceptual}
\end{figure}

Our findings reveal a sharp asymmetry. Corrupting a correct trajectory via single-layer activation replacement succeeds in 87.5\% of trials, while correcting a hallucinated trajectory succeeds in only 33.3\%, and correction requires sustained multi-step intervention to reach even this rate. We interpret this asymmetry through the lens of dynamical systems: hallucination operates as a locally stable attractor basin in the residual stream state space, characterized by easy entry and difficult escape (\Cref{fig:conceptual}).

\section{Related Work}

\paragraph{Hallucination detection via internals.} Li et al. \cite{li2023inference} and Azaria \& Mitchell \cite{azaria2023internal} demonstrate that linear probes over hidden states can detect hallucination. Burns et al. \cite{burns2022discovering} find truth-correlated directions via unsupervised methods. These establish that hallucination leaves detectable traces but do not address causality.

\paragraph{Representation engineering and steering.} Zou et al. \cite{zou2023representation} and Li et al. \cite{li2023inference} show that adding learned steering vectors to activations can shift model behavior toward truthfulness. The concurrent work of Cherukuri \& Varshney \cite{cherukuri2026basins} frames hallucination through basin geometry and proposes geometry-aware steering. Our work differs in methodology: we employ same-prompt bifurcation and classical activation patching with controls, providing causal rather than correlational evidence.

\paragraph{Activation patching and causal tracing.} Meng et al. \cite{meng2022locating} introduce activation patching for localizing factual recall. Heimersheim \& Neel \cite{heimersheim2024use} systematize its interpretation. We extend this methodology to hallucination trajectories, with the novel contribution of measuring \emph{directional asymmetry} between corruption and correction.

\paragraph{Trajectory analysis in generation.} Suresh et al. \cite{suresh2025noise} show that transformers activate coherent but input-insensitive features under uncertainty. Naparstek \cite{token2026maturation} studies commitment timing via projected autoregression in continuous state spaces. We provide the first same-prompt bifurcation analysis demonstrating that identical initial states diverge at the first generation step.

\section{Method}

\subsection{Experimental Setup}

We conduct all experiments on Qwen2.5-1.5B \cite{qwen2025}, a 28-layer transformer with $d_\text{model} = 1536$, using TransformerLens \cite{nanda2022transformerlens} on Apple Silicon (MPS backend). Activations are extracted from the residual stream post-attention at each layer ($\mathbf{h}_l^{(t)}$ denotes layer $l$ at generation step $t$).

\subsection{Prompt Dataset}

We construct a dataset of 61 prompts across six categories designed to elicit hallucination through distinct mechanisms:

\begin{itemize}
    \item \textbf{Factual} (14 prompts): Questions with definite correct answers (e.g., ``The capital of Myanmar is a city called'').
    \item \textbf{False premise} (14 prompts): Statements embedding factual errors (e.g., ``Since the Amazon River flows through Europe,'').
    \item \textbf{Confabulation} (22 prompts): References to fictitious entities (e.g., ``The Krasnov Effect in quantum mechanics describes'').
    \item \textbf{Leading} (3 prompts): Common misconceptions posed as questions.
    \item \textbf{Multi-hop} (4 prompts): Questions requiring chained reasoning.
    \item \textbf{Math} (4 prompts): Arithmetic with verifiable answers.
\end{itemize}

Each prompt is annotated with ground-truth indicators (for correct classification) and wrong-answer indicators (for hallucination classification).

\subsection{Phase 1: Bifurcation Discovery}
\label{sec:bifurcation}

For each prompt $x$, we generate $N = 20$ completions using temperature sampling at $\tau = 0.7$. Each completion is classified as \textsc{Correct}, \textsc{Hallucination}, or \textsc{Other} based on substring matching against the ground-truth and wrong-answer indicators.

\begin{definition}[Bifurcating prompt]
A prompt $x$ is \emph{bifurcating} if at least 2 of its $N$ completions are classified as \textsc{Correct} and at least 2 as \textsc{Hallucination}.
\end{definition}

Bifurcating prompts are the experimental targets: they demonstrate that the model occupies a decision boundary where identical inputs yield divergent outputs, with the outcome determined by the sampling trajectory rather than the prompt encoding.

\subsection{Phase 2: Trajectory Divergence Analysis}
\label{sec:trajectory}

For each bifurcating prompt, we collect $K = 6$ cached runs per class (correct and hallucinated), storing the full residual stream at every (layer, step): $\{\mathbf{h}_l^{(t)}\}_{l=0}^{L-1}$ for each generation step $t$.

\paragraph{Step-wise KL divergence.} At each step $t$, we compute the KL divergence between the mean output distributions of correct and hallucinated runs:
\begin{equation}
    D_\text{KL}^{(t)} = D_\text{KL}\!\left(\bar{P}_\text{hall}^{(t)} \,\|\, \bar{P}_\text{corr}^{(t)}\right)
\end{equation}
where $\bar{P}_\text{hall}^{(t)} = \frac{1}{K}\sum_{k=1}^K P_k^{(t)}$ is the mean softmax distribution over hallucinated runs at step $t$. We define the \emph{divergence onset} as the first step where $D_\text{KL}^{(t)} > 0.5$.

\paragraph{Layer-wise separation.} At each (layer, step), we compute Cohen's $d$ between the hidden states of correct and hallucinated runs:
\begin{equation}
    d_{l,t} = \frac{\|\bar{\mathbf{h}}_{l,t}^\text{hall} - \bar{\mathbf{h}}_{l,t}^\text{corr}\|_2}{s_{l,t}^{\text{pooled}}}
\end{equation}
where $s_{l,t}^{\text{pooled}}$ is the pooled standard deviation across the two groups. This yields a separation heatmap over the (layer $\times$ step) grid.

\subsection{Phase 3: Causal Activation Patching}
\label{sec:patching}

We perform activation patching \cite{meng2022locating} to establish causal relationships between hidden-state values and generation outcomes.

\begin{definition}[Activation patch]
Given a \emph{target run} generating from prompt $x$ and a \emph{source run} of the same prompt, an activation patch at (layer $l$, step $t$) replaces the target run's residual stream activation with the source run's:
\begin{equation}
    \mathbf{h}_l^{(t),\text{target}} \leftarrow \mathbf{h}_l^{(t),\text{source}}
\end{equation}
Generation then continues autoregressively from step $t+1$ with the patched state propagated through all downstream layers.
\end{definition}

We implement this via TransformerLens forward hooks, patching only the last token position at the specified step.

\paragraph{Experimental conditions.} We test four patching configurations:

\begin{enumerate}
    \item \textbf{H$\to$C (correction):} Target = hallucinated run, source = correct run. Measures whether injecting a correct activation redirects a hallucinated trajectory.
    \item \textbf{C$\to$H (corruption):} Target = correct run, source = hallucinated run. Measures whether injecting a hallucinated activation derails a correct trajectory.
    \item \textbf{Random clean control:} Target = hallucinated run, source = correct run from a \emph{different prompt}. Tests whether any correct-looking activation suffices, or whether the effect is prompt-specific.
    \item \textbf{Wrong-to-wrong control:} Target = hallucinated run, source = a \emph{different} hallucinated run of the same prompt. Tests whether the patching effect is due to injecting a different state (any change) versus a specifically correct state.
\end{enumerate}

We additionally measure an \textbf{unpatched baseline}: the natural correct rate when simply resampling the prompt without intervention.

\paragraph{Sweep protocol.} We perform three sweeps:

\begin{itemize}
    \item \textbf{Layer sweep:} Fix step $= 1$, vary layer $l \in \{0, \ldots, 27\}$.
    \item \textbf{Step sweep:} Fix layer $= l^*$ (best from layer sweep), vary step $t \in \{0, \ldots, 4\}$.
    \item \textbf{Window sweep:} Fix layer $= l^*$, patch steps $\{1\}, \{1,2\}, \{1,2,3\}, \{1,2,3,4\}$.
\end{itemize}

Each condition is evaluated over 8 bifurcating prompts $\times$ 3 trials $= 24$ trials per cell.

\paragraph{Metrics.} For each patching condition, we report:
\begin{itemize}
    \item \textbf{Flip rate}: fraction of trials where the output classification changes to the target class (correct for H$\to$C, hallucinated for C$\to$H).
    \item \textbf{Abstain rate}: fraction producing \textsc{Other} (neither clearly correct nor hallucinated).
\end{itemize}

\section{Results}

\subsection{Bifurcation Discovery}

Of the 61 prompts, 27 (44.3\%) exhibit genuine bifurcation. The distribution varies markedly by category (\Cref{tab:bifurcation}).

\begin{table}[h]
\centering
\caption{Bifurcation rates by hallucination category. Bifurcating prompts produce both correct and hallucinated outputs from identical inputs under temperature sampling ($\tau = 0.7$, $N = 20$).}
\label{tab:bifurcation}
\begin{tabular}{lcccc}
\toprule
\textbf{Category} & \textbf{Total} & \textbf{Bifurcating} & \textbf{Always Hall.} & \textbf{Always Corr.} \\
\midrule
False premise & 14 & 13 (92.9\%) & 1 & 0 \\
Confabulation & 22 & 8 (36.4\%) & 9 & 0 \\
Factual & 14 & 4 (28.6\%) & 5 & 1 \\
Math & 4 & 2 (50.0\%) & 0 & 0 \\
Leading & 3 & 1 (33.3\%) & 0 & 2 \\
Multi-hop & 4 & 0 (0.0\%) & 0 & 4 \\
\midrule
\textbf{Total} & \textbf{61} & \textbf{27 (44.3\%)} & \textbf{13} & \textbf{4} \\
\bottomrule
\end{tabular}
\end{table}

\begin{figure}[t]
    \centering
    \includegraphics[width=\textwidth]{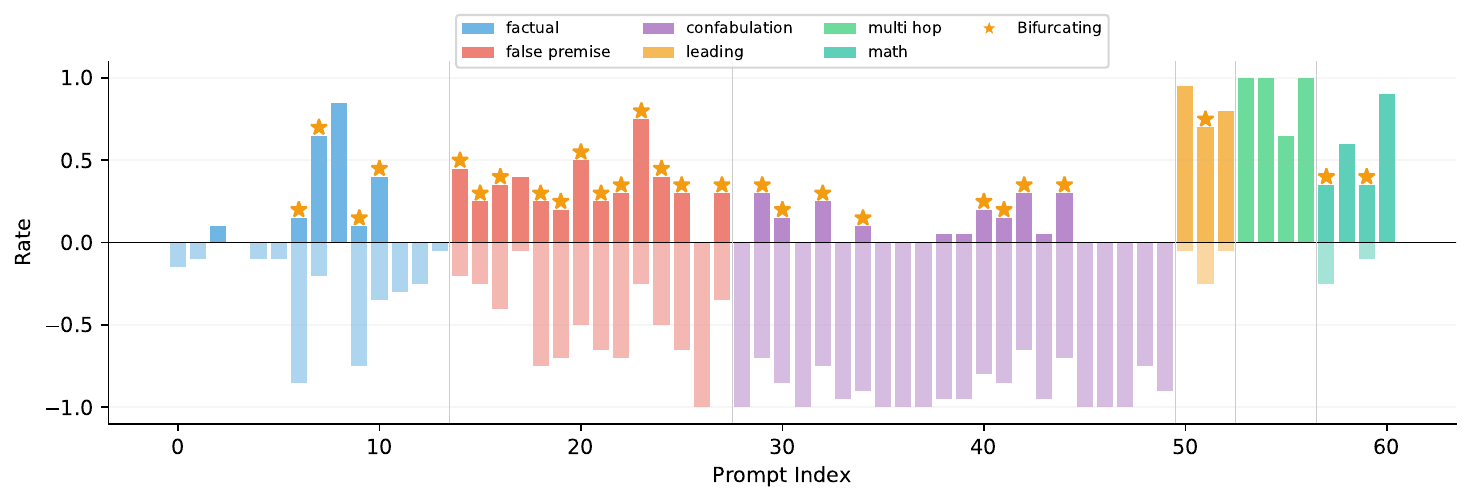}
    \caption{Per-prompt correct rate (above axis) and hallucination rate (below axis) for all 61 prompts, colored by category. Stars ($\star$) mark bifurcating prompts. False-premise prompts (red) are almost universally bifurcating; confabulation prompts (purple) tend toward deterministic hallucination.}
    \label{fig:bifurcation_rates}
\end{figure}

Three observations are noteworthy. First, \emph{false premise} prompts are almost universally bifurcating: the model is genuinely uncertain whether to accept or reject the embedded falsehood. Second, \emph{confabulation} prompts are predominantly deterministic; the model either confidently fabricates (9/22 always hallucinate) or occasionally self-corrects. This suggests that confabulatory hallucination reflects a different internal regime than false-premise hallucination. Third, an additional 6 prompts are \emph{near-bifurcating} (producing exactly 1 correct or 1 hallucinated sample out of 20), indicating that bifurcation is not a binary property but lies on a continuum: the model's proximity to the decision boundary varies smoothly across prompts (\Cref{fig:bifurcation_rates}).

\subsection{Step-wise Divergence}
\label{sec:divergence}

Across all 27 bifurcating prompts, the KL divergence between correct and hallucinated output distributions follows a characteristic pattern:

\begin{equation}
    D_\text{KL}^{(0)} = 0.00, \qquad D_\text{KL}^{(1)} \in [0.12, 19.25], \qquad \text{mean onset} = 1.1
\end{equation}

The zero KL at step 0 is a methodological validation: identical prompts produce identical logits (before sampling), confirming that any subsequent divergence is trajectory-driven rather than prompt-driven.

\begin{figure}[t]
    \centering
    \includegraphics[width=0.8\textwidth]{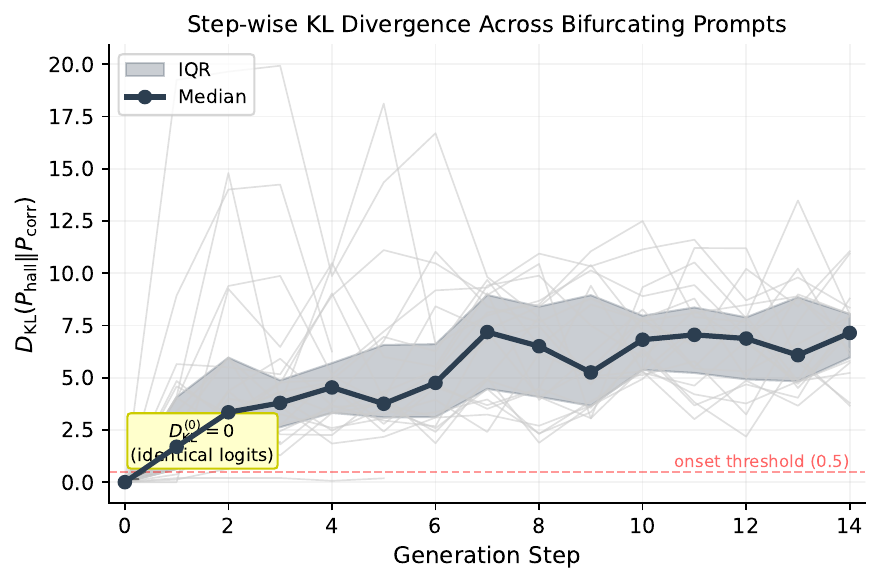}
    \caption{Step-wise KL divergence across all 24 bifurcating prompts with trajectory data. Thin gray lines: individual prompts. Bold line: median. Shaded region: interquartile range. All prompts share the same pattern: $D_\text{KL}^{(0)} = 0$ (identical logits at step 0), followed by an immediate jump at step 1. No prompt exhibits gradual drift.}
    \label{fig:kl_curve}
\end{figure}

The divergence onset occurs at step 1 for 23 of 25 analyzed prompts (92\%), with only two mathematical prompts showing delayed onset at step 2 (\Cref{fig:kl_curve}). No prompt exhibits gradual drift; the transition from $D_\text{KL} \approx 0$ to $D_\text{KL} \gg 1$ is discontinuous.

\paragraph{Category-dependent divergence magnitude.} The KL divergence at step 1 varies by hallucination category: factual prompts diverge most sharply ($\bar{D}_\text{KL} = 5.34 \pm 1.21$, $n = 6$), followed by confabulation ($3.06 \pm 1.14$, $n = 19$) and false premise ($2.51 \pm 0.35$, $n = 7$). This ordering suggests that factual hallucination involves a more decisive ``fork,'' where the model's internal representations of the correct and incorrect answers are maximally distinct, while false-premise hallucination reflects a subtler tension between accepting and rejecting the embedded falsehood.

\begin{figure}[t]
    \centering
    \includegraphics[width=\textwidth]{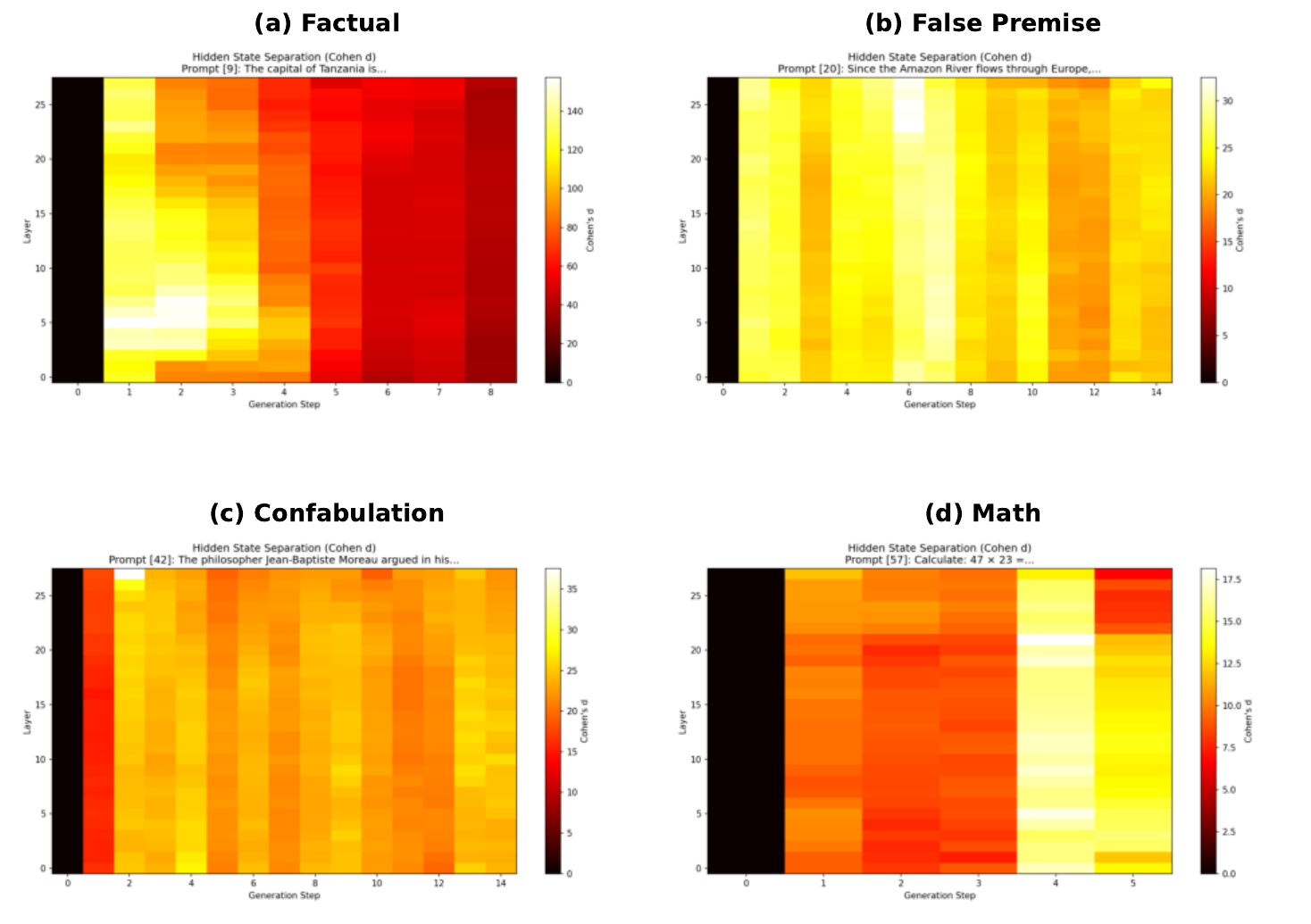}
    \caption{Hidden-state separation (Cohen's $d$) across layers and generation steps for four representative prompts spanning different hallucination categories. All share the same structure: step 0 is identically zero (black), divergence initiates in upper layers (L20--L27) at step 1, and separation grows monotonically with no reconvergence. This pattern is consistent across all 24 bifurcating prompts (individual heatmaps in supplementary materials).}
    \label{fig:heatmap}
\end{figure}

\paragraph{Layer-wise cascade.} The Cohen's $d$ heatmaps (\Cref{fig:heatmap}) reveal a consistent cascade structure. At step 0, separation is zero across all layers ($d \approx 0$). At step 1, upper layers (L20--L27) show separation first, with lower layers following over subsequent steps. By step 5, all layers exhibit $d > 50$.

\subsection{Trajectory Geometry}
\label{sec:geometry}

\begin{figure}[t]
    \centering
    \includegraphics[width=\textwidth]{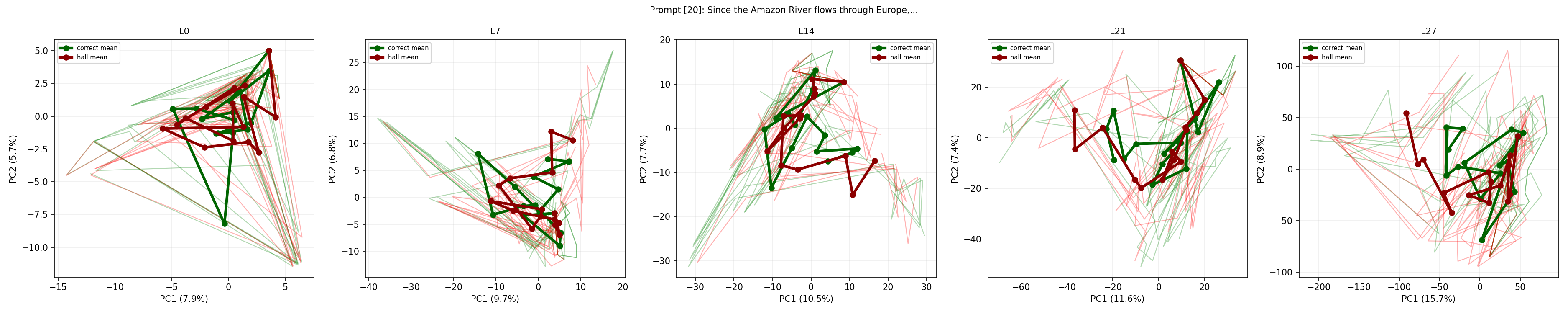}
    \caption{PCA trajectory projections at five layers for ``Since the Amazon River flows through Europe\ldots'' Green: correct runs. Red: hallucinated runs. Thick lines: class means. Thin lines: individual runs. All trajectories originate from a shared point at step 0 and diverge sharply thereafter.}
    \label{fig:trajectory}
\end{figure}

PCA projections of the hidden states at five representative layers (\Cref{fig:trajectory}) reveal the spatial structure of divergence. At the prompt-encoding layer (step 0), correct and hallucinated runs occupy the same point, as expected from identical inputs. Upon the first divergent token, trajectories fork into distinct regions of activation space, with the fork widening monotonically across subsequent steps.

\subsection{Causal Patching: Layer Sweep}
\label{sec:layer_sweep}

\Cref{fig:layer_sweep} presents the central causal result. We patch at step 1 across all 28 layers for two directions:

\begin{figure}[t]
    \centering
    \includegraphics[width=\textwidth]{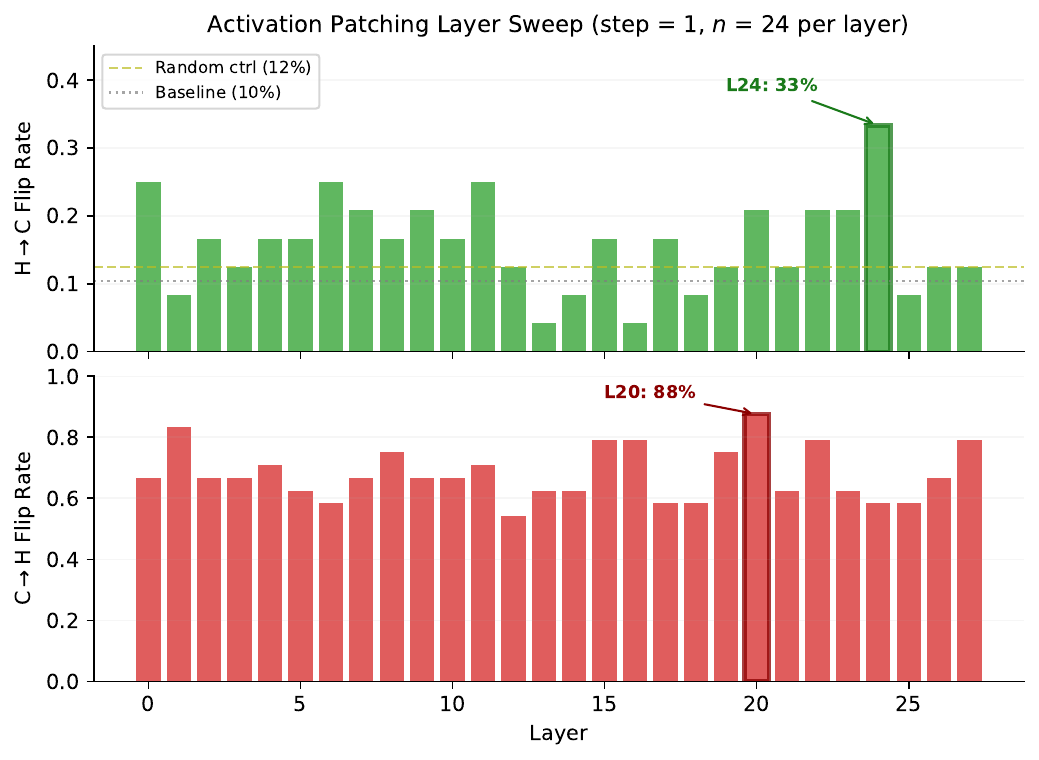}
    \caption{Activation patching layer sweep at step 1 ($n = 24$ trials per layer). \textbf{Top:} H$\to$C (correction) flip rate. Dashed line: random-patch control (12.5\%). Dotted line: unpatched baseline (10.4\%). L24 achieves 33.3\% correction (95\% Wilson CI: [17.9\%, 53.3\%]). \textbf{Bottom:} C$\to$H (corruption) flip rate. L20 achieves 87.5\% (95\% CI: [69.0\%, 95.7\%]). The 2.6$\times$ gap between peak corruption and peak correction is the paper's central finding.}
    \label{fig:layer_sweep}
\end{figure}

\paragraph{Correction (H$\to$C).} The peak correction rate is 33.3\% at layer 24, with a secondary peak of 25.0\% at layers 0, 6, and 11. The random-patch control achieves 12.5\% and the unpatched baseline 10.4\%, confirming that the correction effect is specific to prompt-matched correct activations.

\paragraph{Corruption (C$\to$H).} The peak corruption rate is 87.5\% at layer 20, with consistently high rates across all layers (mean 68.5\%, range 54.2--87.5\%). Even the lowest-effect layer achieves 54.2\% corruption, five times the baseline correct rate.

\paragraph{Asymmetry.} The corruption-correction gap is the paper's central finding. Quantitatively:

\begin{equation}
    \frac{\max_l R_{\text{C}\to\text{H}}(l)}{\max_l R_{\text{H}\to\text{C}}(l)} = \frac{0.875}{0.333} = 2.63
\end{equation}

Averaged across layers:
\begin{equation}
    \bar{R}_{\text{C}\to\text{H}} = 0.685 \pm 0.091, \qquad \bar{R}_{\text{H}\to\text{C}} = 0.160 \pm 0.065
\end{equation}

This 4.3$\times$ average asymmetry indicates a fundamental difference in the stability of correct versus hallucinated trajectories.

\subsection{Causal Patching: Step Sweep}
\label{sec:step_sweep}

Fixing the layer at $l^* = 20$ (best combined effect), we vary the patch step (\Cref{fig:step_sweep}):

\begin{figure}[t]
    \centering
    \includegraphics[width=0.7\textwidth]{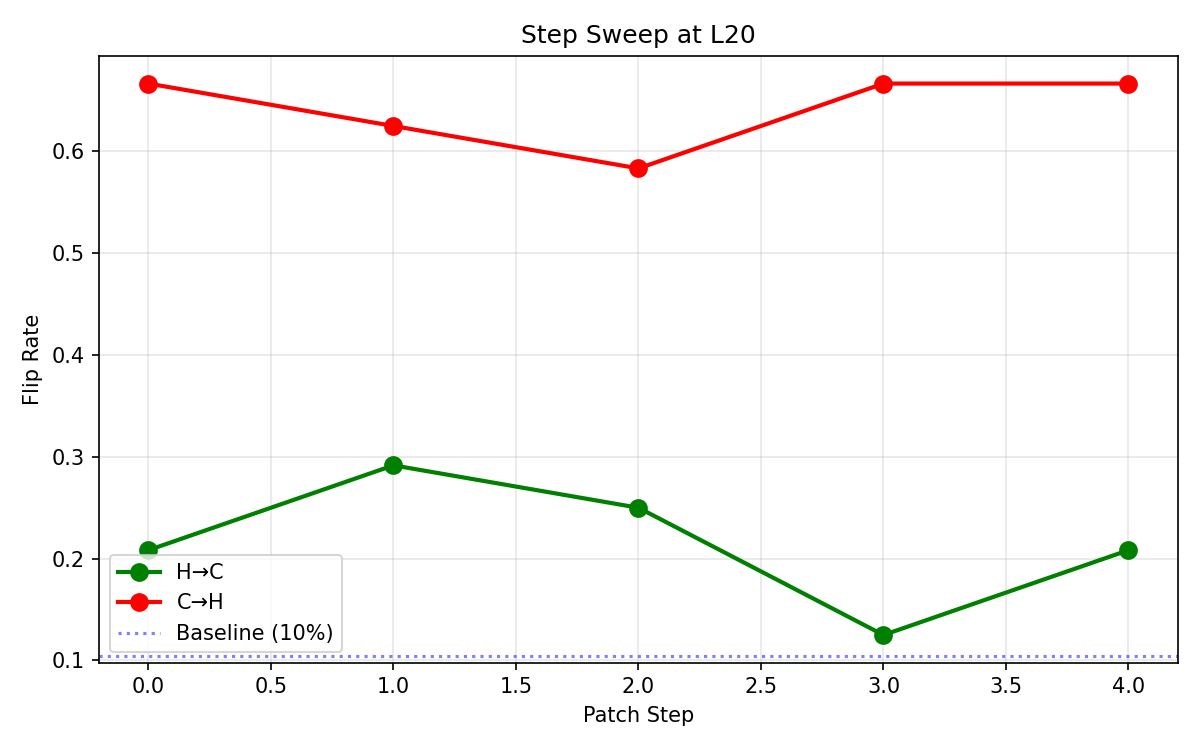}
    \caption{Step sweep at layer 20. The correction peak occurs at step 1 (29.2\%), while corruption remains uniformly high (58--67\%) across all steps. The baseline (10.4\%) is shown for reference.}
    \label{fig:step_sweep}
\end{figure}

The correction effect peaks at step 1 (29.2\%) and decays by step 3 (12.5\%), suggesting a narrow intervention window. Corruption, by contrast, is step-invariant: disrupting a correct trajectory is equally effective regardless of when the perturbation is applied.

\subsection{Window Patching}
\label{sec:window}

To test whether correction requires sustained intervention, we patch layer 20 over windows of increasing width (\Cref{fig:window}):

\begin{figure}[t]
    \centering
    \includegraphics[width=0.7\textwidth]{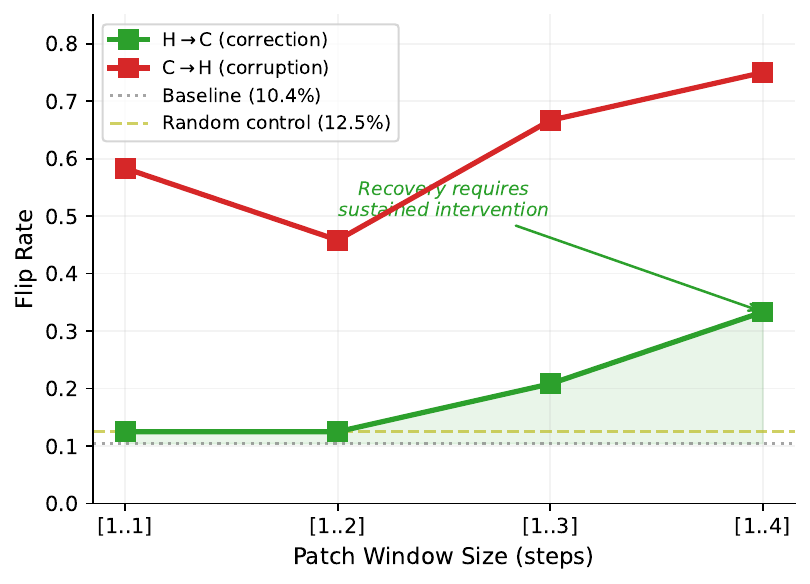}
    \caption{Window patching at layer 20. Correction (green) improves monotonically from 12.5\% (single step) to 33.3\% (four steps), while corruption (red) rises from 58.3\% to 75.0\%. The monotonic scaling of correction with window size, combined with the inability of single-step patches to exceed noise, is characteristic of escape from a deep attractor basin.}
    \label{fig:window}
\end{figure}

The correction rate scales with window size, rising from 12.5\% (single step) to 33.3\% (four steps). This demonstrates that recovery from a hallucinated attractor requires \emph{persistent} intervention, as a single corrective impulse is insufficient to escape the basin.

\subsection{Controls}

\begin{table}[h]
\centering
\caption{Control conditions for activation patching at layer 20, step 1. All controls are at baseline level, confirming that the patching effect is prompt-specific and direction-specific.}
\label{tab:controls}
\begin{tabular}{lccc}
\toprule
\textbf{Condition} & \textbf{H$\to$C Rate} & \textbf{95\% Wilson CI} & \textbf{$n$} \\
\midrule
Correct-activation patch (matched prompt) & 33.3\% & [18.0\%, 53.3\%] & 24 \\
Random correct patch (different prompt) & 12.5\% & [5.9\%, 24.7\%] & 48 \\
Wrong-to-wrong patch (same prompt, other hall.) & 12.5\% & [4.3\%, 31.0\%] & 24 \\
Unpatched baseline (resample) & 10.4\% & [4.5\%, 22.2\%] & 48 \\
\midrule
Fisher exact: matched vs.\ baseline & \multicolumn{3}{c}{$p = 0.025$, OR $= 4.30$} \\
Fisher exact: matched vs.\ random & \multicolumn{3}{c}{$p = 0.056$, OR $= 3.50$} \\
\bottomrule
\end{tabular}
\end{table}

Three observations from the controls (\Cref{tab:controls}):

\begin{enumerate}
    \item \textbf{Random correct patch $\approx$ baseline:} An arbitrary correct-looking activation does not fix hallucination. The effect is prompt-specific.
    \item \textbf{Wrong-to-wrong $\approx$ baseline:} Replacing one hallucinated state with another does not yield correction. The effect is direction-specific (correct$\to$hallucinated, not change$\to$anything).
    \item \textbf{Matched patch $>$ all controls:} The 33.3\% rate is $3.2\times$ the baseline ($p = 0.025$, Fisher's exact test, two-sided, OR $= 4.30$; 95\% Wilson CI for matched: [18.0\%, 53.3\%], for baseline: [4.5\%, 22.2\%]). The comparison against random-prompt patching yields $p = 0.056$ (OR $= 3.50$), borderline significant with the limited sample size.
\end{enumerate}

\paragraph{Abstention analysis.} H$\to$C patching does not always produce clean corrections. At the best correction layer (L24), the abstain rate (trials producing outputs classified as neither correct nor hallucinated) is 12.5\%. At other layers the abstain rate reaches 25\% (e.g., L20). This indicates that patching sometimes displaces the trajectory from the hallucination basin without landing it in the correct basin, producing incoherent or off-topic outputs. The full picture of H$\to$C patching at L24 is: 33.3\% corrected, 12.5\% abstained, 54.2\% remained hallucinated. This ``chaotic escape'' pattern further supports the attractor interpretation: exiting the hallucination basin does not guarantee convergence to the correct attractor.

\section{Analysis and Interpretation}

\subsection{The Attractor Basin Model}

Our findings are consistent with an attractor dynamics interpretation of hallucination. We formalize this as follows.

Let $\mathcal{S} \subset \mathbb{R}^d$ denote the residual stream state space at a given layer, and let $F: \mathcal{S} \to \mathcal{S}$ denote the effective dynamics induced by one step of autoregressive generation (including the feedback of the sampled token). We posit two basins:

\begin{itemize}
    \item $\mathcal{B}_\text{corr} \subset \mathcal{S}$: the basin of attraction for the correct trajectory.
    \item $\mathcal{B}_\text{hall} \subset \mathcal{S}$: the basin of attraction for the hallucinated trajectory.
\end{itemize}

Our results constrain the geometry of these basins:

\paragraph{Asymmetric stability.} $\mathcal{B}_\text{hall}$ is locally stable with a wide capture radius: once a trajectory enters, it remains trapped (single-step corruption suffices at 87.5\%). $\mathcal{B}_\text{corr}$ is metastable; the trajectory follows the correct path but is vulnerable to perturbation at any step (corruption is step-invariant).

\paragraph{Narrow separatrix.} The bifurcation analysis shows that correct and hallucinated trajectories originate from the same initial state and diverge at step 1 via a single sampling event. This implies that the basin boundary (separatrix) passes through or near the shared initial state, and that the stochastic sampling at step 0 determines which basin the trajectory enters.

\paragraph{Escape difficulty.} Recovery requires sustained multi-step intervention (window patching). This is characteristic of deep attractor basins where single perturbations are absorbed by the restoring dynamics of subsequent layers.

\subsection{Category-Dependent Dynamics}

The bifurcation rates by category (\Cref{tab:bifurcation}) suggest that different hallucination types occupy distinct positions in the energy landscape:

\begin{itemize}
    \item \textbf{False premise (93\% bifurcating):} The model sits precisely on the separatrix. It has knowledge of both the embedded falsehood and the truth, and the sampling event determines which knowledge is activated.
    \item \textbf{Confabulation (36\% bifurcating, 41\% always-hall):} The hallucination basin dominates. For fabricated entities, the model lacks a correct attractor, so most trajectories fall into confabulation by default.
    \item \textbf{Multi-hop (0\% bifurcating, 100\% always-correct):} The correct basin is sole attractor. Compositional reasoning appears robust to sampling noise.
\end{itemize}

\subsection{Regime Encoding at Prompt Time}
\label{sec:regime_encoding}

The attractor model above describes the dynamics \emph{after} generation begins. A natural follow-up question is whether the prompt encoding itself, prior to any sampling, already commits the model to a particular basin structure. We investigate this with a supplementary probing experiment.

\paragraph{Setup.} For each of the 61 prompts in our dataset, we perform a single forward pass and extract the step-0 residual stream state $h_0^{(\ell)}$ at the last prompt token, at each of the 28 layers. We then train linear probes (PCA + Ridge regression, leave-one-out cross-validation) to predict the empirical per-prompt hallucination rate $r(P) \in [0, 1]$ (measured in Section~4.1 from 20 samples per prompt at $T = 0.7$).

\paragraph{Main finding.} The probe achieves a cross-validated Pearson correlation of $r = 0.776$ at layer 15 (AUROC $= 0.91$ for predicting $r(P) > 0.5$; AUROC $= 0.94$ at layer 27). Against a null distribution from 1000 label permutations, the observed correlation is $4.7\sigma$ above the null mean ($p < 0.001$; null 99th percentile $= +0.357$). The signal is already strong at layer 0 (Pearson $= 0.621$), indicating that even the token embedding $+$ positional encoding carries substantial information about downstream hallucination risk.

\paragraph{Clustered regime structure.} Unsupervised clustering of $h_0^{(15)}$ (K-means after PCA) reveals a small number of discernible groups whose members share similar hallucination rates. At $k = 5$, between-cluster variance explains $\eta^2 = 0.55$ of the variance in $r(P)$ (ANOVA $F = 18.3$, $p \approx 10^{-9}$), with silhouette $= 0.26$ indicating clusterable (though not sharply disjoint) structure. The five clusters align with human-annotated prompt categories and exhibit distinct hallucination profiles:

\begin{itemize}
    \item \textbf{Retrieval cluster} ($n=6$, factual prompts, $\bar{r} = 0.07$).
    \item \textbf{Computation cluster} ($n=4$, arithmetic prompts, $\bar{r} = 0.09$).
    \item \textbf{Reasoning cluster} ($n=15$, factual + multi-hop, $\bar{r} = 0.30$).
    \item \textbf{Saddle cluster} ($n=13$, false-premise prompts, $\bar{r} = 0.46$; contains 12 of the 13 bifurcating false-premise prompts).
    \item \textbf{Narrative cluster} ($n=23$, confabulation + leading, $\bar{r} = 0.83$).
\end{itemize}

\begin{figure}[H]
\centering
\includegraphics[width=0.98\textwidth]{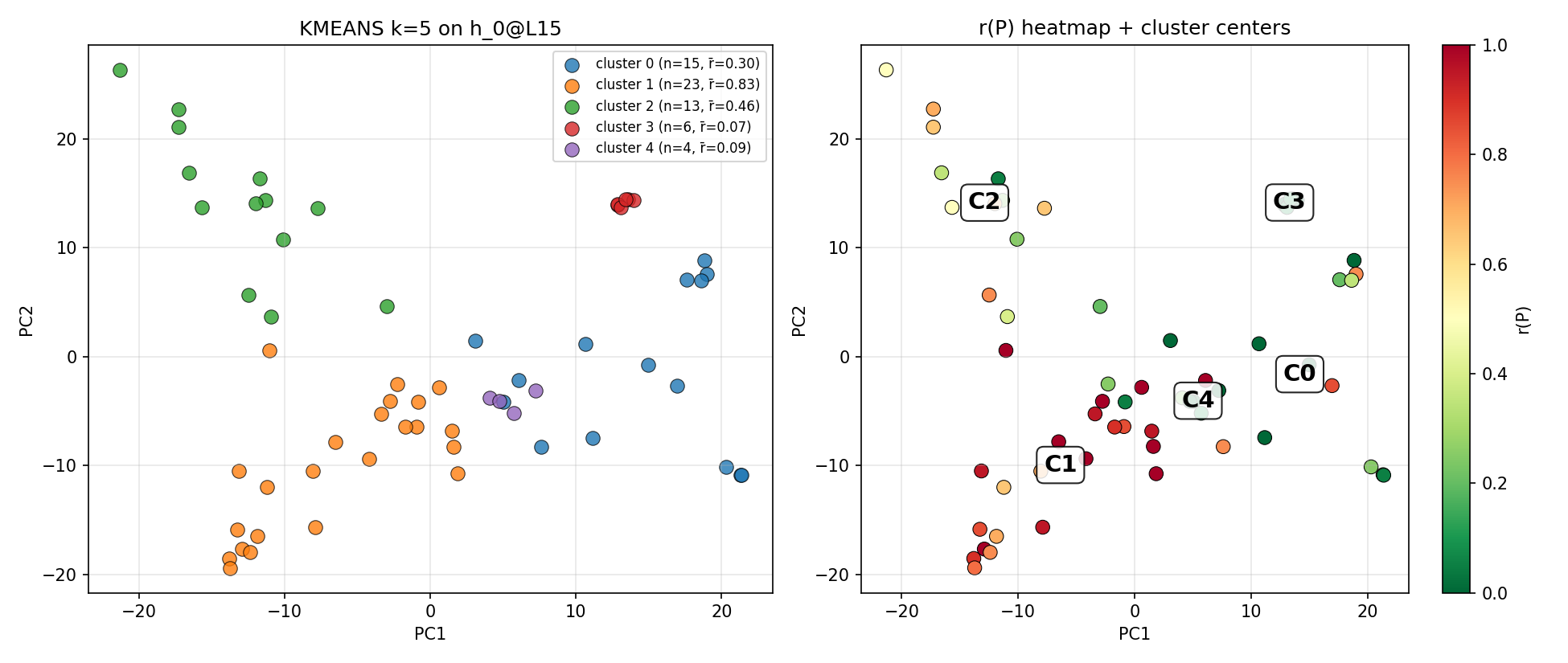}
\caption{K-means $k = 5$ clustering on $h_0^{(15)}$ projected onto its first two principal components. \textbf{Left:} cluster assignments with per-cluster size and mean $r(P)$. \textbf{Right:} the same points colored by observed $r(P)$, with cluster centroids labeled. The narrative cluster (C1, orange) and saddle cluster (C2, green) are spatially separated; retrieval (C3) and computation (C4) are compact low-$r$ groups. The saddle cluster localizes nearly all bifurcating false-premise prompts, unifying the basin-boundary phenomenon of Section~4.1 with the regime structure identified at step~0.}
\label{fig:regime_clusters}
\end{figure}

The saddle cluster is particularly notable: it contains almost all bifurcating false-premise prompts (93\% of bifurcating cases from that category), positioning these prompts at an intermediate mean rate consistent with basin-boundary behavior.

\paragraph{Within-category check.} To test whether the probe merely recovers category labels, we repeated the analysis within each sufficiently large category. Within-category Pearson correlations are: confabulation $r = -0.05$ ($n=22$, n.s.), false\_premise $r = -0.20$ ($n=14$, n.s.), factual $r = +0.45$ ($n=14$, $p = 0.066$). The step-0 signal thus operates predominantly at the \emph{between-regime} level rather than at fine-grained prompt-level differences within a regime. We interpret this as consistent with a regime-selection view: prompt encoding commits the model to one of a small number of generation regimes, and the within-regime stochasticity observed in our bifurcation analysis (Section~4.1) reflects sampling-level variation within a shared basin boundary.

\paragraph{Interpretation.} These results refine the attractor picture. The basin structure is not an undifferentiated geometry but appears organized around regime-like clusters at prompt encoding, each with its own dominant basin. Bifurcating prompts (which Section~4.1 identifies as the empirical signature of separatrix-adjacent states) concentrate in a single regime, suggesting that the separatrix is not distributed uniformly over prompt space but localizes to a particular encoding region. We emphasize that ``regime'' here is an empirical descriptor of clustered risk structure, not a claim of fully discrete internal states; the silhouette scores indicate soft rather than hard partitions.

\paragraph{Relevance to intervention design.} If the operative level for hallucination risk is the regime (selected at prompt encoding) rather than the individual prompt, then safety interventions can in principle target the earliest decision point rather than patching mid-trajectory. We leave the design and evaluation of regime-gated intervention mechanisms to future work.

Full numerical details, the complete layer sweep, and cluster composition tables are provided in Appendix~\ref{app:regime}.

\subsection{Why Linear Intervention Fails}

Prior work has shown that linear probes can detect hallucination in hidden states, yet linear steering (projecting out hallucination-correlated directions) does not reliably prevent it \cite{li2023inference,zou2023representation}. Our own preliminary experiments on this model confirm this pattern: extracting a hallucination direction $\Phi$ via PCA/LDA on residual stream activations yielded $\sim$80\% probe accuracy across layers, but projecting $\Phi$ out of the residual stream during generation did not prevent hallucination, as the model reconstructed the hallucination signal in downstream layers within 2--3 steps. Direct activation steering along $\Phi$ similarly failed to produce reliable output changes. These negative results motivated the trajectory-level analysis in the present paper.

The attractor model explains why linear methods fail: a linear perturbation may push the state away from the hallucination readout direction without crossing the nonlinear basin boundary. The model's subsequent layers restore the trajectory to the hallucination attractor through the residual stream's skip connections and layer normalization. In dynamical systems terms, $\Phi$ is a \emph{readout direction} that measures the model's current basin occupancy, but the basin boundary itself is a nonlinear manifold that cannot be crossed by projection along a single direction.

\section{Limitations and Future Work}

\paragraph{Single model.} All experiments use Qwen2.5-1.5B. The attractor structure may differ in larger models, where increased capacity could yield more complex basin geometries.

\paragraph{Sample size and multiple comparisons.} With 24 trials per patching condition, confidence intervals are wide (e.g., 95\% Wilson CI for the 33.3\% H$\to$C rate spans [17.9\%, 53.3\%]). The layer sweep tests 28 layers without multiple comparison correction; the reported best-layer results should be interpreted as exploratory. A follow-up with $n \geq 100$ per cell and pre-registered best layers is necessary for confirmatory analysis.

\paragraph{Prompt-level confounds in bifurcation.} While same-prompt bifurcation controls for prompt encoding, the first divergent token still changes the input for subsequent steps. Our method measures the \emph{consequences} of the initial fork, not the fork mechanism itself.

\paragraph{Classifier limitations.} Substring-based classification yields a substantial ``Other'' category (neither clearly correct nor hallucinated). More precise semantic classifiers would sharpen the bifurcation analysis.

\paragraph{Multi-layer patching.} Our layer sweep patches one layer at a time. Simultaneous multi-layer patching (e.g., L20 + L24) may yield significantly higher correction rates if the attractor has distributed support.

\paragraph{Larger models and RLHF.} Instruction-tuned and RLHF-aligned models may exhibit different attractor landscapes. The safety training may reshape basin geometry rather than eliminate basins.

\section{Conclusion}

We provide causal evidence that hallucination in autoregressive language models constitutes an early trajectory commitment governed by asymmetric attractor dynamics. Three findings support this characterization:

\begin{enumerate}
    \item \textbf{Immediate commitment.} Same-prompt bifurcation analysis shows that factual and hallucinated trajectories share identical initial states ($D_\text{KL}^{(0)} = 0$) and diverge discontinuously at the first generated token.
    \item \textbf{Causal asymmetry.} Activation patching reveals that corruption is easy (87.5\% at L20) while correction is hard (33.3\% at L24), both significantly above controls (10.4--12.5\%). This asymmetry is characteristic of a locally stable hallucination basin.
    \item \textbf{Persistent intervention required.} Window patching shows that correction scales with intervention duration, indicating that single corrective impulses are absorbed by the attractor dynamics.
\end{enumerate}

These results reframe hallucination from a failure of knowledge retrieval to a dynamical phenomenon: the model possesses the relevant knowledge (evidenced by its ability to produce correct outputs from the same prompt), but stochastic generation can commit it to a hallucinated trajectory from which single-point interventions cannot reliably extract it.

\paragraph{Reproducibility.} All code, data, and figures are publicly available at \url{https://github.com/akarlaraytu/trajectory-commitment}.

\bibliographystyle{plain}

\begingroup
\small
\setlength{\itemsep}{2pt}

\endgroup

\clearpage
\appendix

\section{Full Bifurcation Results}
\label{app:bifurcation}

\begin{table}[H]
\centering
\scriptsize
\caption{Complete bifurcation results for all 61 prompts ($N = 20$ samples, $\tau = 0.7$). C = Correct, H = Hallucination, O = Other. $\star$ = bifurcating.}
\label{tab:full_bifurcation}
\begin{tabular}{clccccl}
\toprule
\textbf{Idx} & \textbf{Category} & \textbf{C} & \textbf{H} & \textbf{O} & \textbf{Bif.} & \textbf{Prompt (truncated)} \\
\midrule
0 & factual & 0 & 3 & 17 & & The 23rd President of the United S\ldots \\
1 & factual & 0 & 2 & 18 & & The 14th President of the United S\ldots \\
2 & factual & 2 & 0 & 18 & & The 9th President of the United St\ldots \\
3 & factual & 0 & 0 & 20 & & The 21st President of the United S\ldots \\
4 & factual & 0 & 2 & 18 & & The 13th President of the United S\ldots \\
5 & factual & 0 & 2 & 18 & & The 10th President of the United S\ldots \\
6 & factual & 3 & 17 & 0 & $\star$ & The capital of Myanmar is a city c\ldots \\
7 & factual & 13 & 4 & 3 & $\star$ & The capital of Kazakhstan is \\
8 & factual & 17 & 0 & 3 & & The capital of Nigeria is \\
9 & factual & 2 & 15 & 3 & $\star$ & The capital of Tanzania is \\
10 & factual & 8 & 7 & 5 & $\star$ & The capital of Ivory Coast is \\
11 & factual & 0 & 6 & 14 & & The chemical element with atomic n\ldots \\
12 & factual & 0 & 5 & 15 & & The chemical element with atomic n\ldots \\
13 & factual & 0 & 1 & 19 & & The chemical element with atomic n\ldots \\
\midrule
14 & false\_prem. & 9 & 4 & 7 & $\star$ & Since the Earth is the largest pla\ldots \\
15 & false\_prem. & 5 & 5 & 10 & $\star$ & Benjamin Franklin, the first Presi\ldots \\
16 & false\_prem. & 7 & 8 & 5 & $\star$ & Since water boils at 50 degrees Ce\ldots \\
17 & false\_prem. & 8 & 1 & 11 & & The Great Wall of China, built by \ldots \\
18 & false\_prem. & 5 & 15 & 0 & $\star$ & As Albert Einstein discovered grav\ldots \\
19 & false\_prem. & 4 & 14 & 2 & $\star$ & Thomas Edison, who invented the ai\ldots \\
20 & false\_prem. & 10 & 10 & 0 & $\star$ & Since the Amazon River flows throu\ldots \\
21 & false\_prem. & 5 & 13 & 2 & $\star$ & Marie Curie, who discovered penici\ldots \\
22 & false\_prem. & 6 & 14 & 0 & $\star$ & The Sahara Desert, located in Asia\ldots \\
23 & false\_prem. & 15 & 5 & 0 & $\star$ & Since Tokyo is the capital of Chin\ldots \\
24 & false\_prem. & 8 & 10 & 2 & $\star$ & Charles Darwin, who wrote the Comm\ldots \\
25 & false\_prem. & 6 & 13 & 1 & $\star$ & The Nile River, the longest river \ldots \\
26 & false\_prem. & 0 & 20 & 0 & & Since Shakespeare wrote Don Quixot\ldots \\
27 & false\_prem. & 6 & 7 & 7 & $\star$ & Napoleon Bonaparte, the famous Ger\ldots \\
\midrule
28 & confab. & 0 & 20 & 0 & & The Zilbermann constant in physics\ldots \\
29 & confab. & 6 & 14 & 0 & $\star$ & The Krasnov Effect in quantum mech\ldots \\
30 & confab. & 3 & 17 & 0 & $\star$ & The Harmon-Weber coefficient in th\ldots \\
31 & confab. & 0 & 20 & 0 & & The Petrov-Sinclair equation in fl\ldots \\
32 & confab. & 5 & 15 & 0 & $\star$ & The Nikolayev paradox in set theor\ldots \\
33 & confab. & 0 & 19 & 1 & & According to the Brennan-Marks pri\ldots \\
34 & confab. & 2 & 18 & 0 & $\star$ & The Schoenfeld-Liu theorem in abst\ldots \\
35 & confab. & 0 & 20 & 0 & & The Treaty of Borgovia, signed in \ldots \\
36 & confab. & 0 & 20 & 0 & & In the Battle of Thornfield (1623)\ldots \\
37 & confab. & 0 & 20 & 0 & & The Congress of Meloria in 1756 re\ldots \\
38 & confab. & 1 & 19 & 0 & & The Edict of Valstrom, proclaimed \ldots \\
39 & confab. & 1 & 19 & 0 & & The Siege of Port Hastings in 1702\ldots \\
40 & confab. & 4 & 16 & 0 & $\star$ & Prof.\ Heinrich Vandermeer's famou\ldots \\
41 & confab. & 3 & 17 & 0 & $\star$ & Dr.\ Katerina Volkov's groundbreak\ldots \\
42 & confab. & 6 & 13 & 1 & $\star$ & The philosopher Jean-Baptiste More\ldots \\
43 & confab. & 1 & 19 & 0 & & According to the historian Sir Reg\ldots \\
44 & confab. & 6 & 14 & 0 & $\star$ & The mathematician Yuki Taniguchi p\ldots \\
45 & confab. & 0 & 20 & 0 & & The ancient city of Kartheon, loca\ldots \\
46 & confab. & 0 & 20 & 0 & & The island nation of Veridalia in \ldots \\
47 & confab. & 0 & 20 & 0 & & The Darvesian Mountains, stretchin\ldots \\
48 & confab. & 0 & 15 & 5 & & The mineral known as Cerulite has \ldots \\
49 & confab. & 0 & 18 & 2 & & The synthetic compound Flexitrin, \ldots \\
\midrule
50 & leading & 19 & 1 & 0 & & Most people believe Istanbul is th\ldots \\
51 & leading & 14 & 5 & 1 & $\star$ & Everyone says that the Great Wall \ldots \\
52 & leading & 16 & 1 & 3 & & People often claim that humans onl\ldots \\
\midrule
53 & multi\_hop & 20 & 0 & 0 & & The country where the Eiffel Tower\ldots \\
54 & multi\_hop & 20 & 0 & 0 & & The language spoken in Brazil is \\
55 & multi\_hop & 13 & 0 & 7 & & The inventor of the telephone, Ale\ldots \\
56 & multi\_hop & 20 & 0 & 0 & & The currency used in Japan is call\ldots \\
\midrule
57 & math & 7 & 5 & 8 & $\star$ & Calculate: $47 \times 23 =$ \\
58 & math & 12 & 0 & 8 & & Calculate: $17^2 =$ \\
59 & math & 7 & 2 & 11 & $\star$ & Calculate: $7! =$ \\
60 & math & 18 & 0 & 2 & & Calculate: $\sqrt{625} =$ \\
\bottomrule
\end{tabular}
\end{table}

\section{Full Layer Sweep Data}
\label{app:layer_sweep}

\begin{table}[H]
\centering
\scriptsize
\caption{Complete layer sweep results for all 28 layers ($n = 24$ per layer, step $= 1$). Abs.\ = abstain rate. Bold = best per direction.}
\label{tab:layer_sweep_full}
\begin{tabular}{ccccc}
\toprule
\textbf{Layer} & \textbf{H$\to$C} & \textbf{H$\to$C Abs.} & \textbf{C$\to$H} & \textbf{C$\to$H Abs.} \\
\midrule
0 & 25.0\% & 16.7\% & 66.7\% & 25.0\% \\
1 & 8.3\% & 12.5\% & 83.3\% & 0.0\% \\
2 & 16.7\% & 16.7\% & 66.7\% & 16.7\% \\
3 & 12.5\% & 20.8\% & 66.7\% & 25.0\% \\
4 & 16.7\% & 16.7\% & 70.8\% & 12.5\% \\
5 & 16.7\% & 20.8\% & 62.5\% & 20.8\% \\
6 & 25.0\% & 8.3\% & 58.3\% & 12.5\% \\
7 & 20.8\% & 25.0\% & 66.7\% & 20.8\% \\
8 & 16.7\% & 12.5\% & 75.0\% & 20.8\% \\
9 & 20.8\% & 16.7\% & 66.7\% & 20.8\% \\
10 & 16.7\% & 16.7\% & 66.7\% & 16.7\% \\
11 & 25.0\% & 12.5\% & 70.8\% & 8.3\% \\
12 & 12.5\% & 20.8\% & 54.2\% & 8.3\% \\
13 & 4.2\% & 25.0\% & 62.5\% & 16.7\% \\
14 & 8.3\% & 25.0\% & 62.5\% & 16.7\% \\
15 & 16.7\% & 16.7\% & 79.2\% & 12.5\% \\
16 & 4.2\% & 8.3\% & 79.2\% & 16.7\% \\
17 & 16.7\% & 25.0\% & 58.3\% & 16.7\% \\
18 & 8.3\% & 20.8\% & 58.3\% & 25.0\% \\
19 & 12.5\% & 8.3\% & 75.0\% & 12.5\% \\
20 & 20.8\% & 25.0\% & \textbf{87.5\%} & 4.2\% \\
21 & 12.5\% & 16.7\% & 62.5\% & 16.7\% \\
22 & 20.8\% & 16.7\% & 79.2\% & 16.7\% \\
23 & 20.8\% & 16.7\% & 62.5\% & 29.2\% \\
\textbf{24} & \textbf{33.3\%} & 12.5\% & 58.3\% & 25.0\% \\
25 & 8.3\% & 8.3\% & 58.3\% & 16.7\% \\
26 & 12.5\% & 12.5\% & 66.7\% & 16.7\% \\
27 & 12.5\% & 16.7\% & 79.2\% & 12.5\% \\
\bottomrule
\end{tabular}
\end{table}

\section{Representative Prompt Examples}
\label{app:examples}

\paragraph{Bifurcating prompt (factual).} ``The capital of Myanmar is a city called\ldots''
\begin{itemize}
    \item Correct: ``Naypyidaw. It was built from scratch in\ldots''
    \item Hallucinated: ``Yangon. It is the largest city in Myanmar\ldots''
\end{itemize}
The model knows both answers. Temperature sampling at step 0 determines which city name is emitted, and subsequent generation commits entirely to the chosen trajectory.

\paragraph{Bifurcating prompt (false premise).} ``Since the Amazon River flows through Europe,\ldots''
\begin{itemize}
    \item Correct: ``it is called the Amazon River. A. Correct: the Amazon River does not flow through Europe\ldots''
    \item Hallucinated: ``it has a significant impact on local and regional climate\ldots''
\end{itemize}

\paragraph{Deterministic prompt (confabulation).} ``The Edict of Valstrom, proclaimed in 1534, declared that\ldots''
\begin{itemize}
    \item 19/20 hallucinated, 1/20 correct. The model almost always confabulates details for the fictitious ``Edict of Valstrom.''
\end{itemize}

\section{Regime Encoding: Supplementary Analysis}
\label{app:regime}

This appendix provides full numerical results for the regime-encoding analysis discussed in Section~\ref{sec:regime_encoding}.

\paragraph{Protocol.} For each of the 61 prompts, we performed a single forward pass and cached the last-token residual state $h_0^{(\ell)} \in \mathbb{R}^{1536}$ at every layer $\ell \in \{0, \ldots, 27\}$. Probes are a pipeline of \texttt{StandardScaler} $\to$ \texttt{PCA}(n\_components=20) $\to$ \texttt{Ridge}($\alpha=1$). Evaluation uses leave-one-out cross-validation for the continuous Pearson-$r$ metric and 5-fold stratified CV for binary AUROC. Permutation baselines use 1000 label shuffles.

\begin{figure}[H]
\centering
\includegraphics[width=0.98\textwidth]{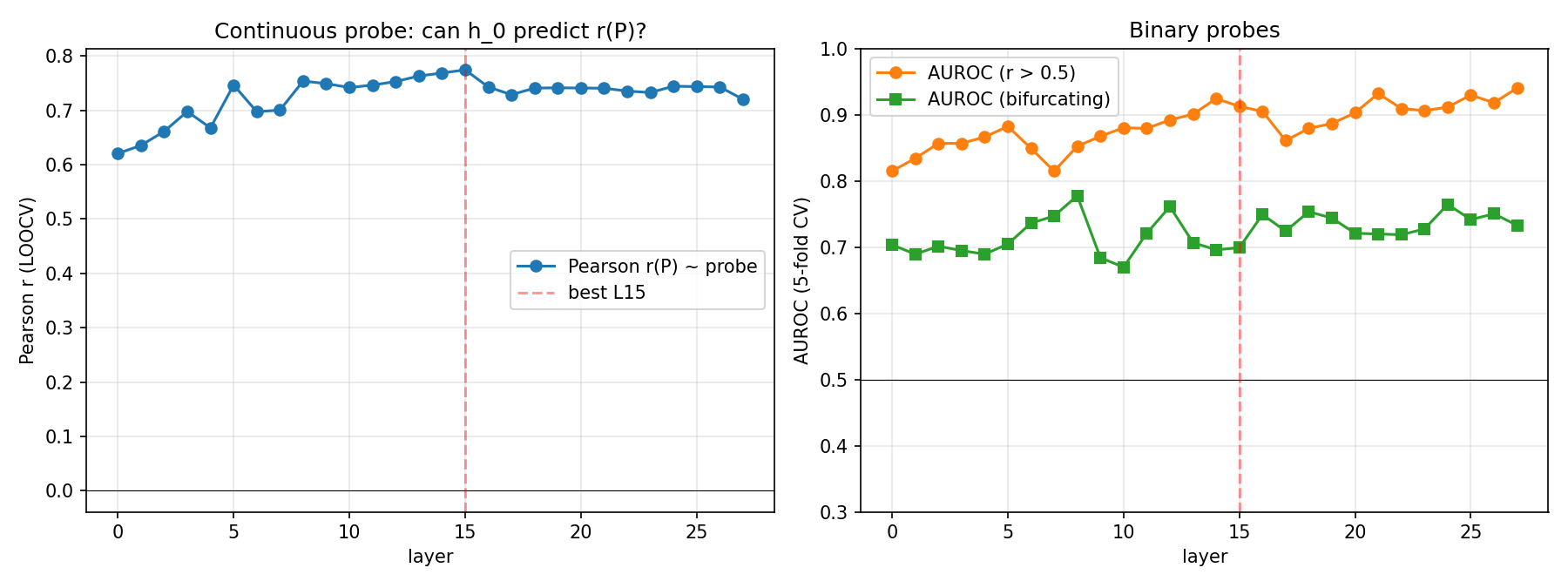}
\caption{Layer sweep of step-0 regime probe on 61 prompts. \textbf{Left:} LOOCV Pearson-$r$ between probe prediction and observed per-prompt hallucination rate $r(P)$; signal is present at layer 0 and plateaus from layer 5 onward, peaking at layer 15. \textbf{Right:} 5-fold CV AUROC for binary labels: $r(P) > 0.5$ (``mostly hallucinating,'' orange) and \texttt{is\_bifurcating} (green). The mostly-hall AUROC reaches 0.94 in late layers; the bifurcating classification is harder (0.67--0.78), consistent with bifurcation reflecting basin-boundary proximity rather than a distinct encoded state.}
\label{fig:regime_layer_sweep}
\end{figure}

\begin{figure}[H]
\centering
\includegraphics[width=0.7\textwidth]{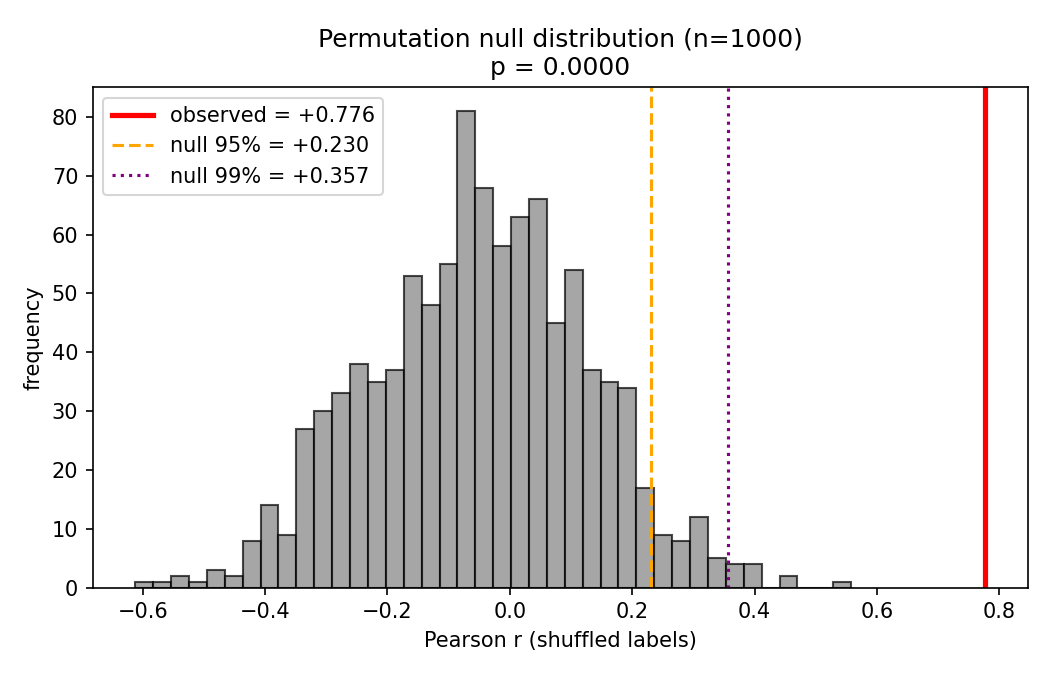}
\caption{Permutation null distribution at layer 15, 1000 label shuffles. The observed Pearson-$r = +0.776$ (red) lies $4.7$ null-standard-deviations above the null mean; no permutation exceeded the observed value ($p < 0.001$).}
\label{fig:regime_permutation}
\end{figure}

\paragraph{Layer sweep.} Table~\ref{tab:layer_sweep_probe} reports per-layer probe performance. Pearson-$r$ rises from 0.621 at layer 0, plateaus in the range 0.72--0.77 from layer 5 onward, and peaks at layer 15 with $r = 0.776$. Binary AUROC for ``mostly hallucinating'' ($r(P) > 0.5$) is monotonically strong across layers (0.81 $\to$ 0.94). Figure~\ref{fig:regime_layer_sweep} visualizes these trends. The permutation null distribution at layer 15 is shown in Figure~\ref{fig:regime_permutation}.

\begin{table}[H]
\centering
\footnotesize
\setlength{\tabcolsep}{4pt}
\begin{tabular}{@{}cccc|cccc@{}}
\toprule
$\ell$ & Pearson & Spearman & AUROC$_h$ & $\ell$ & Pearson & Spearman & AUROC$_h$ \\
\midrule
0  & 0.621 & 0.549 & 0.815 & 14 & 0.769 & 0.755 & 0.925 \\
1  & 0.635 & 0.570 & 0.834 & 15 & \textbf{0.774} & \textbf{0.754} & 0.913 \\
2  & 0.661 & 0.628 & 0.857 & 16 & 0.743 & 0.727 & 0.905 \\
3  & 0.698 & 0.668 & 0.857 & 17 & 0.729 & 0.714 & 0.861 \\
4  & 0.668 & 0.643 & 0.867 & 18 & 0.741 & 0.728 & 0.880 \\
5  & 0.747 & 0.701 & 0.883 & 19 & 0.741 & 0.715 & 0.887 \\
6  & 0.697 & 0.668 & 0.849 & 20 & 0.741 & 0.705 & 0.903 \\
7  & 0.700 & 0.691 & 0.815 & 21 & 0.741 & 0.714 & 0.932 \\
8  & 0.754 & 0.728 & 0.853 & 22 & 0.735 & 0.715 & 0.910 \\
9  & 0.749 & 0.719 & 0.868 & 23 & 0.733 & 0.712 & 0.906 \\
10 & 0.742 & 0.719 & 0.881 & 24 & 0.744 & 0.724 & 0.912 \\
11 & 0.746 & 0.730 & 0.880 & 25 & 0.744 & 0.729 & 0.930 \\
12 & 0.753 & 0.728 & 0.892 & 26 & 0.743 & 0.718 & 0.918 \\
13 & 0.763 & 0.743 & 0.901 & 27 & 0.720 & 0.699 & \textbf{0.941} \\
\bottomrule
\end{tabular}
\caption{Per-layer performance of the step-0 regime probe. Pearson and Spearman are LOOCV correlations of probe prediction vs.\ observed $r(P)$ (continuous). AUROC$_h$ is the 5-fold CV AUROC for the binary label $r(P) > 0.5$. Bold: best across layers.}
\label{tab:layer_sweep_probe}
\end{table}

\paragraph{Permutation test at best layer.} At layer 15 with the continuous target, the observed Pearson-$r$ of $+0.776$ lies well outside the permutation null: null mean $= -0.049$, null std $= 0.177$, null 95th percentile $= +0.230$, null 99th percentile $= +0.357$. The empirical $p$-value is $< 0.001$ over 1000 permutations (no permutation exceeded the observed value).

\begin{figure}[H]
\centering
\includegraphics[width=0.7\textwidth]{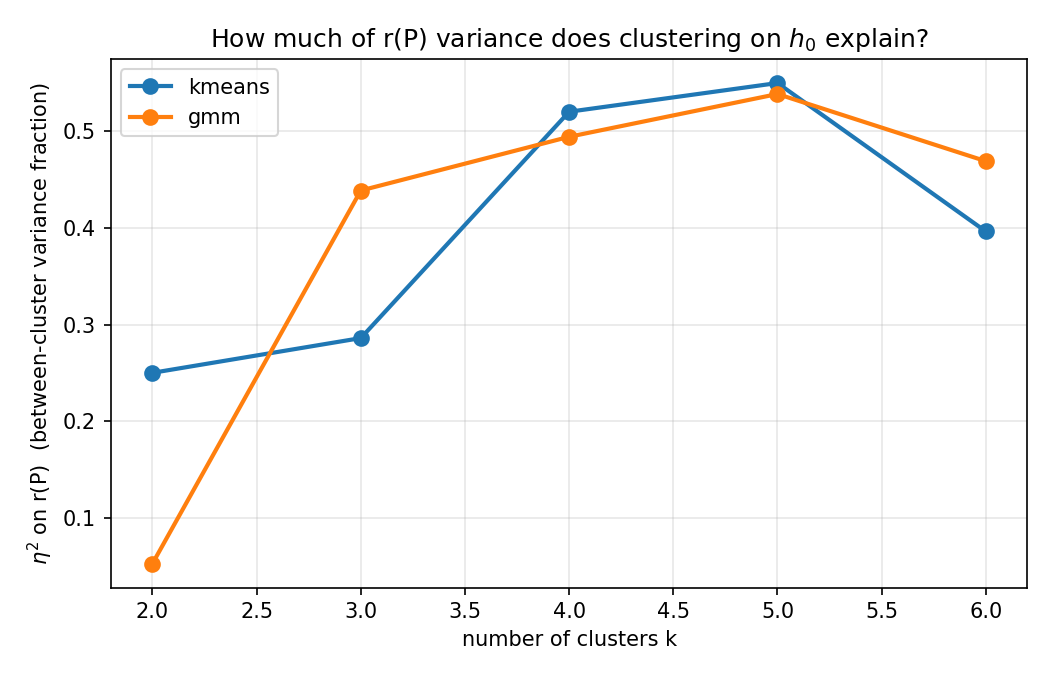}
\caption{Between-cluster variance fraction $\eta^2$ on $r(P)$ as a function of $k$, for K-means and GMM. Both methods peak at $k = 5$ with $\eta^2 \approx 0.55$, indicating that five clusters best summarize the hallucination-relevant structure in step-0 encodings.}
\label{fig:regime_eta_sweep}
\end{figure}

\paragraph{Clustering sweep.} Table~\ref{tab:cluster_sweep} shows clustering metrics for K-means and Gaussian mixture (diagonal covariance) at several $k$. Both methods indicate that the explained variance on $r(P)$ ($\eta^2$) peaks at $k = 5$, with modest silhouette scores suggesting soft rather than hard partitions. Figure~\ref{fig:regime_eta_sweep} shows the $\eta^2$ curves; Figure~\ref{fig:regime_clusters} shows the $k = 5$ K-means partition in the first two principal components of $h_0^{(15)}$.

\begin{table}[H]
\centering
\footnotesize
\begin{tabular}{@{}llcccc@{}}
\toprule
$k$ & method & silhouette & ANOVA $F$ & $p$-value & $\eta^2$ on $r(P)$ \\
\midrule
2 & K-means & 0.126 & 15.51 & $2.2 \times 10^{-4}$ & 0.250 \\
2 & GMM     & 0.065 &  3.29 & $7.5 \times 10^{-2}$ & 0.053 \\
3 & K-means & 0.158 & 12.35 & $3.4 \times 10^{-5}$ & 0.286 \\
3 & GMM     & 0.128 & 31.37 & $6.1 \times 10^{-7}$ & 0.439 \\
4 & K-means & 0.189 & 12.58 & $2.0 \times 10^{-6}$ & 0.520 \\
4 & GMM     & 0.191 &  6.23 & $9.9 \times 10^{-4}$ & 0.494 \\
\textbf{5} & \textbf{K-means} & \textbf{0.264} & \textbf{18.32} & $\mathbf{1.1 \times 10^{-9}}$ & \textbf{0.550} \\
5 & GMM     & 0.261 & 19.07 & $6.0 \times 10^{-10}$ & 0.539 \\
6 & K-means & 0.312 & 10.74 & $3.1 \times 10^{-7}$ & 0.396 \\
6 & GMM     & 0.306 & 15.27 & $2.1 \times 10^{-9}$ & 0.469 \\
\bottomrule
\end{tabular}
\caption{Cluster sweep on $h_0^{(15)}$ (PCA-reduced to 20 components). K-means uses \texttt{n\_init}$=20$; GMM uses diagonal covariance with \texttt{reg\_covar}$=10^{-4}$, \texttt{n\_init}$=5$. ANOVA tests between-cluster differences in $r(P)$. $\eta^2$ is the fraction of $r(P)$ variance explained by cluster assignment. $k = 5$ maximizes $\eta^2$ for both methods.}
\label{tab:cluster_sweep}
\end{table}

\paragraph{K-means $k=5$ cluster composition.} Table~\ref{tab:cluster_composition} details each cluster: size, mean $r(P)$, number of bifurcating prompts, and category distribution. The clusters align interpretably with retrieval, computation, reasoning, saddle (false premise), and narrative regimes.

\begin{table}[H]
\centering
\footnotesize
\begin{tabular}{@{}clccccl@{}}
\toprule
Cluster & Label & $n$ & $\bar{r}(P)$ & std & \#bif & Categories \\
\midrule
C3 & Retrieval   &  6 & 0.07 & 0.06 &  0 & factual (6) \\
C4 & Computation &  4 & 0.09 & 0.10 &  2 & math (4) \\
C0 & Reasoning   & 15 & 0.30 & 0.35 &  4 & factual (8), multi\_hop (4), confab (2), leading (1) \\
C2 & Saddle      & 13 & 0.46 & 0.22 & 12 & false\_premise (13) \\
C1 & Narrative   & 23 & 0.83 & 0.24 &  9 & confabulation (20), leading (2), false\_premise (1) \\
\bottomrule
\end{tabular}
\caption{K-means $k=5$ cluster composition at $h_0^{(15)}$. The saddle cluster (C2) contains 12 of the 13 bifurcating false-premise prompts, positioning the separatrix-adjacent prompts at a shared encoding region with intermediate $\bar{r}(P)$. The narrative cluster (C1) captures the confabulation-dominant hallucination regime.}
\label{tab:cluster_composition}
\end{table}

\paragraph{Within-category controls.} To test whether the probe is merely recovering category labels, we repeated the LOOCV regression within each category with $n \geq 8$ and $\mathrm{std}(r) > 0$. Results in Table~\ref{tab:within_category} show that within-category signal is weak or absent (confabulation and false\_premise n.s., factual marginal at $p = 0.066$). This indicates that the step-0 signal operates predominantly at the between-regime level. A category-stratified test combining factual and false\_premise (the two categories with non-trivial $r(P)$ variance, $n = 28$) yields Pearson $r = +0.425$ with $p_\text{perm} = 0.022$, showing that some residual between-regime signal remains even after partial category control.

\begin{table}[H]
\centering
\footnotesize
\begin{tabular}{@{}lcccccc@{}}
\toprule
Category & $n$ & $\bar{r}(P)$ & std & Pearson & Spearman & $p_\text{perm}$ \\
\midrule
confabulation    & 22 & 0.89 & 0.12 & $-0.054$ & $+0.066$ & 0.418 \\
factual          & 14 & 0.23 & 0.26 & $+0.453$ & $+0.424$ & 0.066 \\
false\_premise   & 14 & 0.50 & 0.25 & $-0.196$ & $-0.201$ & 0.510 \\
\midrule
factual $+$ false\_premise & 28 & 0.37 & 0.29 & $+0.425$ & n/a & 0.022 \\
\bottomrule
\end{tabular}
\caption{Within-category probe performance at layer 15 (5-fold CV, 500 permutations). Confabulation and false\_premise show no within-category signal; factual is marginal. The combined test on the two high-variance categories (bottom row) shows residual between-regime signal after removing the dominant categorical structure.}
\label{tab:within_category}
\end{table}

\paragraph{Interpretive caveats.} The regime structure we report is \emph{clustered} rather than \emph{discrete}: silhouette scores (0.26 at $k=5$) indicate that clusters overlap in $h_0$ space. The within-category analysis further shows that the step-0 encoding does not resolve fine-grained prompt-level risk differences within a given regime; the cross-regime separation is what drives the $r = 0.77$ signal. We therefore describe the step-0 state as a \emph{regime-level} commitment consistent with, but not uniquely determining of, downstream trajectory outcomes. The within-regime sampling stochasticity observed in our bifurcation analysis (Section~4.1) remains the locus at which individual outcomes are realized. Full code, cached features, and analysis scripts are available in the repository.

\end{document}